# Human Skin Detection Using RGB, HSV and YCbCr Color Models

S. Kolkur[1], D. Kalbande[2], P. Shimpi[2], C. Bapat[2], and J. Jatakia[2]

[1] Department of Computer Engineering, Thadomal Shahani Engineering College, Bandra,Mumbai, India

[2] Department of Computer Engineering, Sardar Patel Institute of Technology, Andheri,Mumbai, India

{ kolkur.seema@gmail.com; drkalbande@spit.ac.in; prajwalshimpi@gmail.com; chai.bapat@gmail.com; jatakiajanvi12@gmail.com}

**Abstract.** Human Skin detection deals with the recognition of skin-colored pixels and regions in a given image. Skin color is often used in human skin detection because it is invariant to orientation and size and is fast to process. A new human skin detection algorithm is proposed in this paper. The three main parameters for recognizing a skin pixel are RGB (Red, Green, Blue), HSV (Hue, Saturation, Value) and YCbCr (Luminance, Chrominance) color models. The objective of proposed algorithm is to improve the recognition of skin pixels in given images. The algorithm not only considers individual ranges of the three color parameters but also takes into account combinational ranges which provide greater accuracy in recognizing the skin area in a given image.

## 1 Introduction

Skin detection is the process of finding skin-colored pixels and regions in an image or a video. This process is typically used as a preprocessing step to find regions that potentially have human faces and limbs in images [2]. Skin image recognition is used in a wide range of image processing applications like face recognition, skin disease detection, gesture tracking and human-computer interaction [1]. The primary key for skin recognition from an image is the skin color. But color cannot be the only deciding factor due to the variation in skin tone according to different races. Other factors such as the light conditions also affect the results. Therefore, the skin tone is often combined with other cues like texture and edge features. This is achieved by breaking down the image into individual pixels and classifying them into skin colored and non-skin colored [1]. One simple method is to check if each skin pixel falls into a defined color range or values in some coordinates of a color space. There are many skin color spaces like RGB, HSV, YCbCr, YIQ, YUV, etc. that are used for skin color segmentation [1]. We have proposed a new threshold based on the combination of RGB, HSV and YCbCr values. The following factors should be considered for determining the threshold range:

1) Effect of illumination depending on the surroundings.
2) Individual characteristics such as age, sex and body parts.
3) Varying skin tone with respect to different races.
4) Other factors such as background colors, shadows and motion blur.

The skin detection is influenced by the parameters like Brightness, Contrast, Transparency, Illumination, and Saturation. The detection is normally optimized by taking into consideration combinations of the mentioned parameters in their ideal ranges.

**ATLANTIS PRESS**

## 2  Literature Review

In today's fast paced life, where personal health care has taken a back seat and lowest priority due to ever-growing hustle for earning more and staying ahead of competition, the significance of health can hardly be over-stated. At such crucial junctures, if technology can join hands with health sector, humanity will be blessed. In rural India, just like in any other developing country, ignorance towards personal health care is so rampant that skin diseases often go unnoticed and overlooked. The uneducated Indian rural masses wouldn't even know they have a skin disease until it reaches the last stage (or the most critical and dangerous phase which is often incurable).

Thus by logic, an intervention done by technology in the primitive or early stages of disease contraction would greatly help the diagnosis and avert the fatalities. Such a prognosis needs accurate detection of human skin.

Skin detection techniques can be broadly classified as pixel-based techniques or region-based techniques. In the pixel-based skin detection, each pixel is classified as either skin or non-skin pixel individually depending on certain conditions. The skin detection based on color values is pixel-based. In region-based skin detection technique, spatial relationship of pixels is considered to define some area from given image as skin region. Initial skin region is grown bigger by adding more pixels based on its neighbors properties [6].

Using machine learning based on available data sets, a classifier can be trained to differentiate the image pixel by pixel (a skin pixel from a non-skin pixel). This can greatly help in setting a range of values which are valid for concluding that the pixel is a skin pixel. Then a set of area from the image is recognized as a skin image, using RGB (Red Green Blue), HSV (Hue Saturation Value) and YCbCr.

This paper uses a threshold based methodology to detect whether an image is a skin image or not. It attempts to give a constructive and feasible solution to skin disease detection problem by implementing the different color models on the skin images. It formulates a range for RGB, HSV and YCbCr models which other papers have not ascertained. These ranges attempt to distinguish the skin pixels from the non-skin pixels. Most of the research work in this area highlights the different methodologies that can be used for image recognition; different color models. However after a comparative study of strengths and weaknesses of these models; combination of RGB, HSV and YCbCr seem to fit for the purpose of recognizing skin images.

## 3  Color Spaces

Color space is a mathematical model to represent color information as three or four different color components. Different color spaces (models) are used for different applications such as computer graphics, image processing, TV broadcasting, and computer vision. Different color space is available for the skin detection. They are: RGB based color space (RGB, normalized RGB), Hue Based color space (HSI, HSV, and HSL), Luminance based color space (YCBCr, YIQ, and YUV)[5]. These models are explained subsequently in next sections. Color space selection is the primary process in skin color modeling and further for classification. One or more color spaces can give an optimal threshold value for detection of pixels of skin in a given image. The choice of appropriate color space is often determined by the skin detection methodology and the application. We use the following color spaces for recognizing skin pixels.

### 3.1  Red, Green, and Blue (RGB) Color Model

RGB color space is widely used and is normally the default color space for storing and representing digital images. We can get any other color space from a linear or non-linear transformation of RGB [1]. The RGB color space is the color space used by computers, graphics cards and monitors or LCDs. As shown in fig.1 it consists of three components, red, green and blue, the primary colors. Any color can be obtained by mixing the three base colors. Depending on how much is taken from each base color, any color can be created. Reversing this technique, a specific color can be broken down into its red, blue and green components as shown in equation 1 to equation 3 [1]. These values can be used to find out similar colored pixels from the image. [7] explains skin color detection based on RGB color space. Normalized RGB is a representation that is easily obtained from the RGB values by a simple normalization procedure [1].

A remarkable property of this representation is that for matte surfaces, while ignoring ambient light, normalized RGB is invariant (under certain assumptions) to changes of surface orientation relatively to the light source [4].

$$r = \frac{R}{R+G+B} \tag{1}$$

$$g = \frac{G}{R+G+B} \tag{2}$$

$$b = \frac{B}{R+G+B} \tag{3}$$

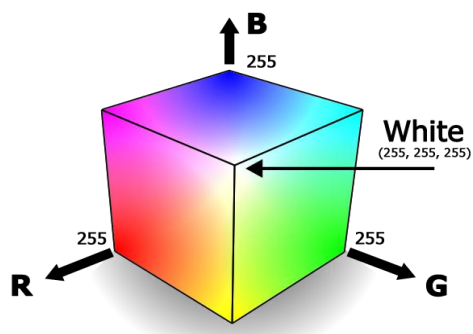

**Fig. 1.** RGB Color Model

## 3.2 YCbCr (Luminance, Chrominance) Color Model

YCbCr is an encoded non-linear RGB signal, commonly used by European television studios and for image compression work. As shown in fig. 2 color is represented by luma (which is luminance computed from non-linear RGB) constructed as a weighted sum of RGB values [4]. YCbCr is a commonly used color space in digital video domain. Because the representation makes it easy to get rid of some redundant color information, it is used in image and video compression standards like JPEG, MPEG1, MPEG2 and MPEG4. The transformation simplicity and explicit separation of luminance and chrominance components makes YCbCr color space [3]. In this format, luminance information is stored as a single component (Y), and chrominance information is stored as two color-difference components (Cb and Cr). Cb represents the difference between the blue component and reference value. Cr represents the difference between the red component and a reference value. YCbCr values can be obtained from RGB color space according to eq. 4 to eq. 6. [7][8][9] uses YCbCr space for skin detection.

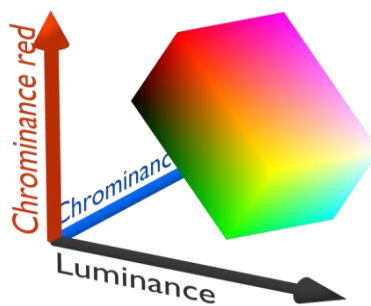

**Fig. 2.** YCbCr Color Model

$Y = 0.299R + 0.287G + 0.11B$                 -------------------eq. 4

$Cr = R - Y$                 -------------------eq. 5

$Cb = B - Y$                 -------------------eq. 6

### *3.3 Hue Saturation Value (HSV) Color Model*

The HSV color space is more intuitive to how people experience color than the RGB color space. As hue (H) varies from 0 to 1.0, the corresponding colors vary from red, through yellow, green, cyan, blue, and magenta, back to red. As saturation(S) varies from 0 to 1.0, the corresponding colors (hues) vary from unsaturated (shades of gray) to fully saturated (no white component). As value (V), or brightness, varies from 0 to 1.0, the corresponding colors become increasingly brighter. The hue component in HSV is in the range 0° to 360° angle all lying around a hexagon as shown figure 3 [3].

With RGB the color will have values like (0.5, 0.5, 0.25), whereas for HSV it will be (30°, √3/4, 0.5). HSV is best used when a user is selecting a color interactively It is usually much easier for a user to get to a desired color as compared to using RGB [3]. [9][11] explain use of HSV color space for skin detection.

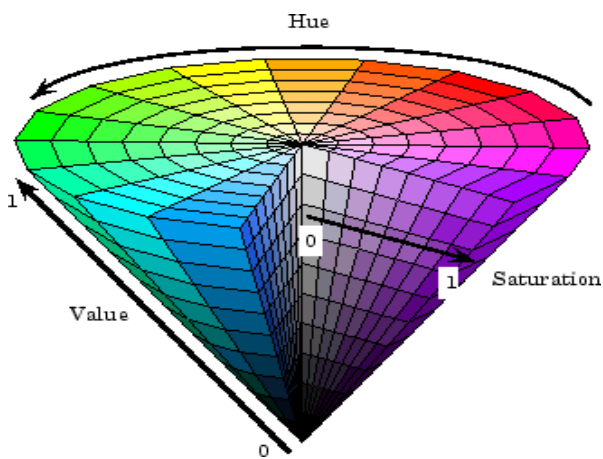

**Fig. 3.**HSV Color Model

## 4 Proposed Skin Detection Algorithm

The proposed algorithm converts the entire image in a two dimensional matrix in which the column and row size is defined by the width and height of the image respectively. Once the image is divided, each entry consists of a pixel of the image. The ARGB color of that particular pixel is determined. The ARGB value retrieved from the image for each pixel is a 32-bit value. Hence to extract each sub-value i.e. red, green, blue and alpha we right shift this value by 24 bit in order to get the value of alpha. The alpha channel is normally used as an opacity channel. If apixel has a value of 0% in its alpha channel, it is fully transparent (and, thus, invisible), whereas a value of 100% in the alpha channel gives a fully opaque pixel (traditional digital images). Similarly, for red right we shift by 16 bits, for green right shift by 8 bits. The remaining value is of blue color.

Bitwise AND operation with 0xff was applied on these calculated values in order to extract only the bits corresponding to that particular color. The above entire procedure is applied to each and every pixel of the image. In order to make the recognition more precise the ARGB value is converted to HSV as well as YCbCr value using conversion factors and built-in functions. The HSV, YCbCr and ARGB value of each pixel is compared to the standard values of a skin pixel and decision is made whether the pixel is a skin pixel or not depending on whether the values lie in a range of predefined threshold values for each parameter.

The ranges for a skin pixel in different color spaces used by our algorithm are as follows:

0.0 <= H <= 50.0 and 0.23 <= S <= 0.68 and

R > 95 and G > 40 and B > 20 and R > G and R > B

and | R - G | > 15 and A > 15

*OR*

R > 95 and G > 40 and B > 20 and R > G and R > B

and | R - G | > 15 and A > 15 and Cr > 135 and

Cb > 85 and Y > 80 and Cr <= (1.5862*Cb)+20 and

Cr>=(0.3448*Cb)+76.2069 and

Cr >= (-4.5652*Cb)+234.5652 and

Cr <= (-1.15*Cb)+301.75 and

Cr <= (-2.2857*Cb)+432.85nothing

(H : Hue ; S: Saturation ; R : Red ; B: Blue ; G : Green ; Cr, Cb : Chrominance components ; Y : luminance component ) Figure 4 shows flowchart that illustrates  steps of the algorithm.

## 5   Experimental Results

*Pratheepan* dataset for human skin detection is used as baseline for comparing results [11]. The images in this dataset are downloaded randomly from Google for human skin detection research. These images are captured with a range of different cameras using different colour enhancement and under different illuminations. The dataset also contains Ground Truth images for sample images in dataset. Fig 5 shows results obtained on some of the images in the dataset. Each diagram shows original image, ground truth image and resultant image from our algorithm. Table 1 shows accuracy calculations on the images shown in fig 5 using following definitions [14]. True positive (TP) represents number of Skin pixels correctly identified as skin, True negative (TN) is number of Non-skin pixel correctly identified as non-skin, False positive (FP) is Non-skin pixel incorrectly identified as skin and False negative (FN) –Skin pixel incorrectly identified as non-skin. Precision and Accuracy is calculated using equations 7 and 8 respectively. Precision of 89.33% and accuracy of 94.43% was obtained on a subset of images from this set.

$$Precision = \frac{TP}{TP+FP} \qquad \text{-------------------eq. 7}$$

$$Accuracy = \frac{TP+TN}{(TP+TN+FP+FN)} \qquad \text{-------------------eq. 8}$$

Table 1. Accuracy Calculations

| Sr. No. | Total no of Pixels | Skin pixels detected (Our algo.) | Skin pixels in GT image | Nonskin Pixels detected (Our algo.) | Nonskin pixels in GT image | True Positive | False Positive | True Negative | False Negative | Precision | Accuracy |
|---|---|---|---|---|---|---|---|---|---|---|---|
| 1 | 196608 | 54885 | 55836 | 140772 | 141723 | 54885 | 0 | 140772 | 951 | 100 | 99.5 |
| 2 | 176418 | 29828 | 23089 | 153329 | 146590 | 23089 | 6739 | 146590 | 0 | 77.4 | 96.1 |
| 3 | 114400 | 20497 | 21128 | 93272 | 93903 | 20497 | 0 | 93272 | 631 | 100 | 99.4 |
| 4 | 108600 | 51191 | 49420 | 59180 | 57409 | 49420 | 1771 | 57409 | 0 | 96.5 | 98.3 |
| 5 | 50000 | 19328 | 18926 | 31074 | 30672 | 18926 | 402 | 30672 | 0 | 97.9 | 99.1 |
| 6 | 128000 | 72237 | 47359 | 80641 | 55763 | 47359 | 24878 | 55763 | 0 | 65.5 | 80.55 |

Additionally some sample images were collected from internet [12]. Fig. 6 shows the results obtained from the algorithm on sample images. The bar chart in fig. 7 represents the number of skin pixels detected in the three different color spaces RGB , HSV & YCbCr for three images respectively 6a)-6c). All three color spaces are almost equally contributing in the process of skin pixel identification. The algorithm is implemented in JAVA.

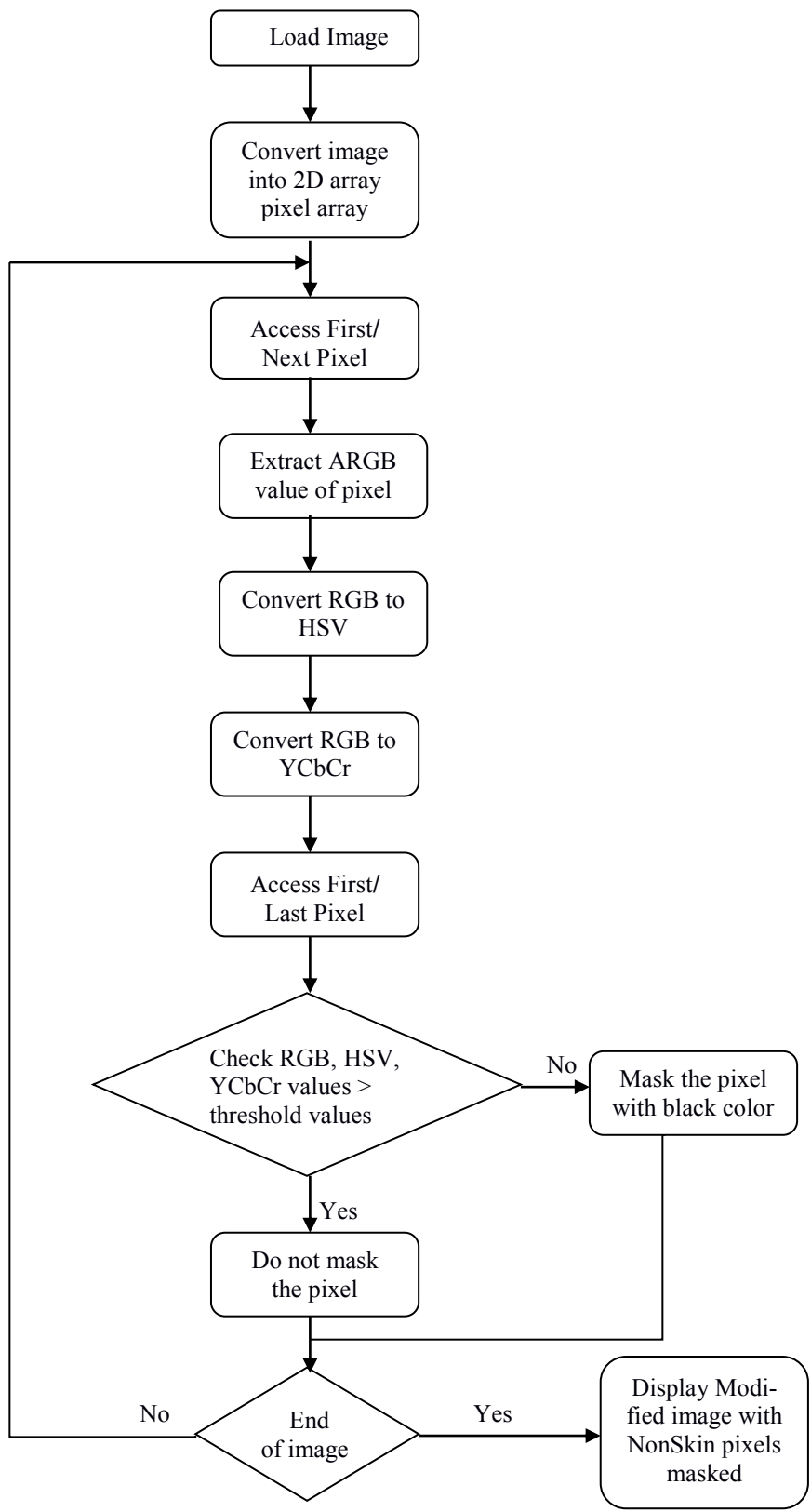

Fig. 4 Flowchart of the proposed system

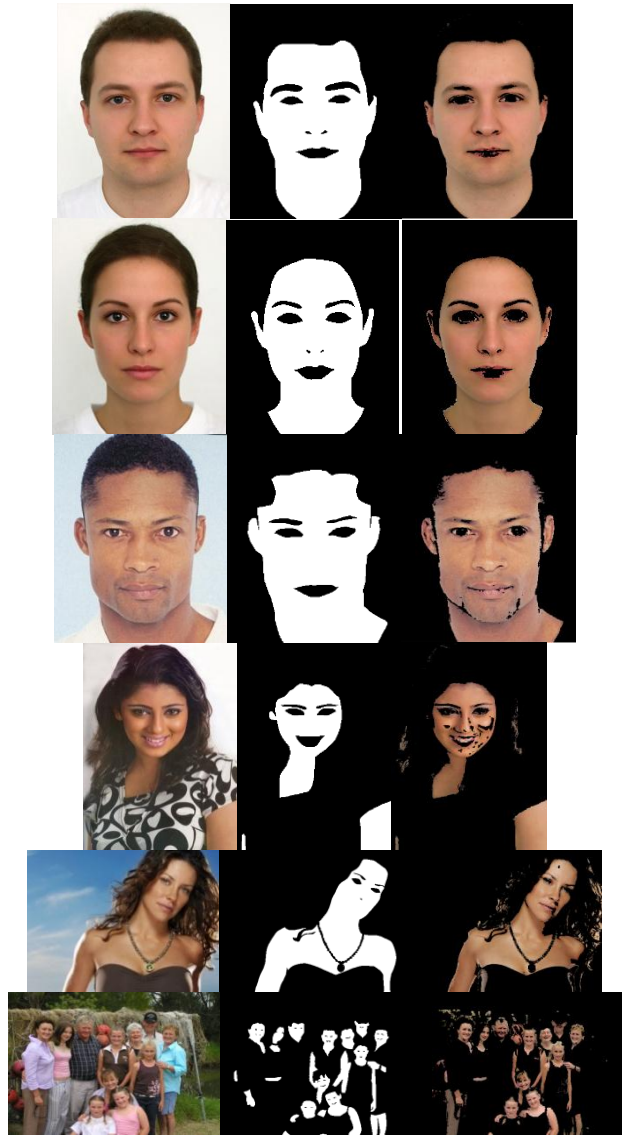

**Fig. 5.** Experimental results on Pratheepan Dataset

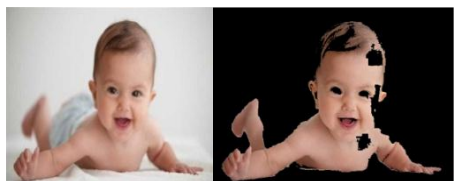

(a)

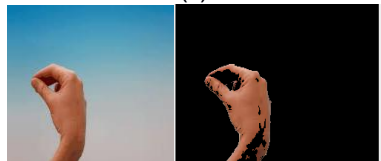

(b)

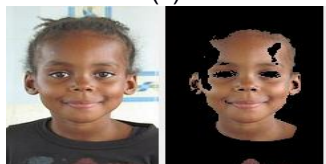

(c)

**Fig. 6.** Experimental results on sample images.

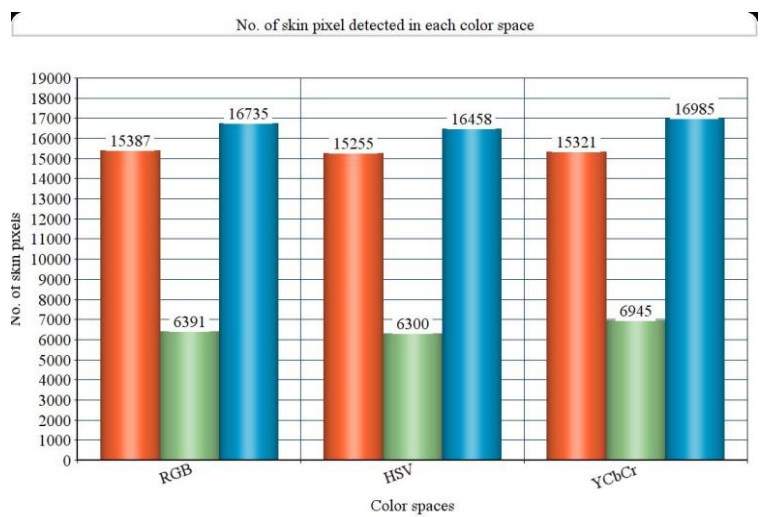

**Fig. 7.** Bar chart showing no. of Skin Pixels

## 6 Conclusion

This paper demonstrates a threshold based algorithm which recognizes skin image using the RGB-HSV-YCbCr model. The algorithm is capable of processing images of different light conditions such as brightness etc. Our algorithm gives promising results in terms of precision and accuracy when compared with baseline dataset as seen in fig. 6. The future scope of this algorithm is to detect face, hand as well as hand gestures which can be used for security purpose, aid for physically challenged (deaf) individuals or for skin disease detection.